\documentclass[conference]{IEEEtran}
\IEEEoverridecommandlockouts

\usepackage[hidelinks]{hyperref}
\usepackage[cmex10]{amsmath}
\usepackage{amssymb,amsfonts}
\usepackage{dblfloatfix}

\usepackage[ruled,vlined]{algorithm2e}
\usepackage{graphicx}
\graphicspath{{Figures/PDF/}{Figures/PNG/}}

\usepackage{booktabs}
\usepackage{siunitx}
\usepackage[numbers,compress]{natbib}
\usepackage{texnames}
\usepackage{bm,bbm}
\usepackage{orcidlink}
\usepackage{multirow}
\usepackage{tabularx}
\usepackage{booktabs}
\usepackage{array}
\usepackage{float}
\usepackage{enumitem}
\usepackage{subcaption}
\usepackage{orcidlink}
\usepackage{hyperref}
\usepackage[font=scriptsize]{caption}

\newcolumntype{C}[1]{>{\centering\arraybackslash}m{#1}}
\newcolumntype{Y}{>{\raggedright\arraybackslash}X}

\def\BibTeX{{\rm B\kern-.05em{\sc i\kern-.025em b}\kern-.08em%
    T\kern-.1667em\lower.7ex\hbox{E}\kern-.125emX}}

\begin{document}

\title{\uppercase{MVT: MASK-GROUNDED VISION-LANGUAGE MODELS FOR TAXONOMY-ALIGNED LAND-COVER TAGGING}

}

\author{%
\IEEEauthorblockN{%
Siyi Chen$^{1,2,*}$\orcidlink{0009-0003-5929-4285},
Kai Wang$^{3,*}$\orcidlink{0009-0000-4751-9553},
Weicong Pang$^{4,*}$\orcidlink{0009-0007-8208-6723},
Ruiming Yang$^{4}$\orcidlink{0009-0000-6016-059X},
Ziru Chen$^{1}$\orcidlink{0009-0009-1732-3651} \\
Renjun Gao$^{5}$\orcidlink{0009-0005-0028-428X},
Alexis Kai Hon Lau$^{1}$\orcidlink{0000-0003-3802-828X},
Dasa Gu$^{1,\dagger}$\orcidlink{0000-0002-5663-1675},
Chenchen Zhang$^{1,\dagger}$\orcidlink{0009-0005-7446-1621},
Cheng Li$^{1,\dagger,\clubsuit}$\orcidlink{0009-0009-4322-7591}\\
}

\\
$^1$ HKUST
$^2$ JHU
$^3$ CUHKSZ
$^4$ NUS
$^5$ MUST \\
{\tt\small clieo@connect.ust.hk \quad dasagu@ust.hk \quad czhangej@connect.ust.hk}\\
\small{* Equal contribution \quad $\dagger$ Corresponding authors \quad $\clubsuit$ Project Leader}
}

\maketitle

\begin{abstract}
Land-cover understanding in remote sensing increasingly demands class-agnostic systems that generalize across datasets while remaining spatially precise and interpretable. We study a geometry-first discovery-and-interpretation setting under domain shift, where candidate regions are delineated class-agnostically and supervision avoids lexical class names via anonymized identifiers. Complementary to open-set recognition and open-world learning, we focus on coupling class-agnostic mask evidence with taxonomy-grounded scene interpretation, rather than unknown rejection or continual class expansion. We propose MVT, a three-stage framework that (i) extracts boundary-faithful region masks using SAM2 with domain adaptation, (ii) performs mask-grounded semantic tagging and scene description generation via dual-step LoRA fine-tuning of multimodal LLMs, and (iii) evaluates outputs with LLM-as-judge scoring calibrated by stratified expert ratings. On cross-dataset segmentation transfer (train on OpenEarthMap, evaluate on LoveDA), domain-adapted SAM2 improves mask quality; meanwhile, dual-step MLLM fine-tuning yields more accurate taxonomy-aligned tags and more informative mask-grounded scene descriptions. The project is available at \url{https://charlescsyyy.github.io/MVT} 
\end{abstract}


\section{Introduction}\label{sec:intro}
Earth observation (EO) is entering a big-data regime, increasing the need for scalable land-cover understanding from remote sensing imagery~\cite{tang2021design,gui2024remote}. 
Yet most pipelines remain closed-set: they assume a fixed taxonomy and degrade under domain shift and emerging or rare land covers~\cite{song2025synthetic,zhang2023remotesensingobjectdetection,yu2024exploringfoundationmodelsremote}. 
This motivates \emph{open-world} remote sensing, where systems should generalize beyond a dataset-specific label set while remaining useful for mapping and monitoring~\cite{wei2025wordsentencelargescalemultiinstance}.

For practical mapping, outputs must be both spatially precise and interpretable: pixel-accurate regions provide geometric evidence, while standardized taxonomy tags enable consistent reporting. 
However, existing open-set methods often decouple these requirements by (i) rejecting ``unknown'' without providing semantic interpretation~\cite{sodano2024open,zheng2024open}, 
(ii) localizing coarsely (e.g., bounding boxes)~\cite{joseph2021openworldobjectdetection}, or 
(iii) relying on predefined vocabulary sets that constrain naming under true novelty~\cite{zhu2024surveyopenvocabularydetectionsegmentation}.

\begin{figure*}[t]
  \centering
  \includegraphics[width=1\textwidth]{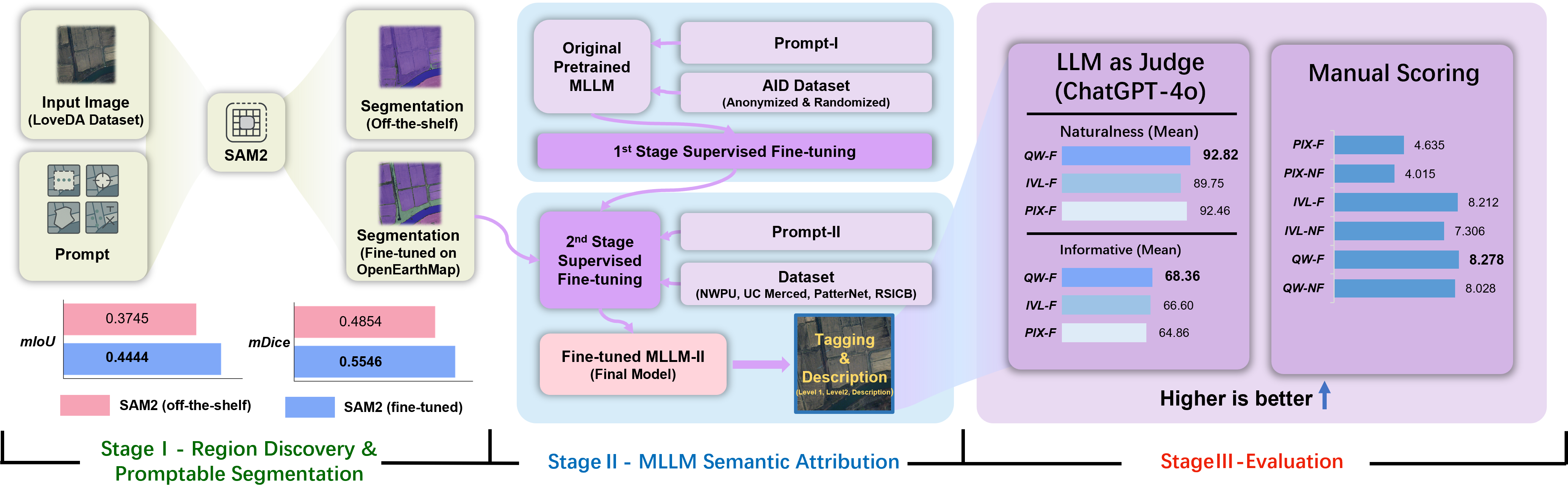}
  \caption{Overview of the proposed MVT framework. The architecture includes the segmentation phase, the two-step fine-tuning of the MLLMs, and the evaluation phase.}
  \label{fig:pipeline}
\end{figure*}

We propose \textbf{MVT}, a geometry-first framework that couples class-agnostic mask discovery with taxonomy-aligned tile-level interpretation under domain shift (Fig.~\ref{fig:pipeline}). 
MVT uses a promptable segmenter (SAM2) to extract boundary-faithful masks as structured evidence, then adapts MLLMs via a lexical-label-free two-step LoRA schedule with mask cues injected as grounding inputs. 
Finally, we evaluate generated descriptions with an LLM-as-judge protocol calibrated by stratified expert ratings~\cite{openai2024gpt4ocard}.

Our contributions are: (1) a class-agnostic discovery-and-interpretation setting for remote sensing under domain shift; 
(2) a mask-grounded, lexical-label-free MLLM tuning strategy that produces taxonomy-aligned tags and grounded descriptions; and 
(3) a scalable evaluation protocol combining LLM judging with expert calibration.

\section{Related Work} \label{sec:related}
\subsection{class-agnostic Perception and Promptable Segmentation}
Remote-sensing semantic segmentation has evolved from CNN encoder--decoder architectures to stronger context-aggregation and transformer-based models that better capture multi-scale cues and preserve boundaries~\cite{long2015fullyconvolutionalnetworkssemantic,badrinarayanan2016segnetdeepconvolutionalencoderdecoder,wang2024improved,9324289,Wang_2022,10347750}. Beyond architectural advances, Remote Sensing (RS) specific efforts integrate multi-sensor fusion, structured refinement, shape priors, and domain generalization to mitigate cross-sensor and seasonal shifts~\cite{rs17040590}. Despite progress on benchmarks such as OpenEarthMap and LoveDA~\cite{xia2022openearthmapbenchmarkdatasetglobal,wang2022lovedaremotesensinglandcover}, most pipelines remain closed-set and rely on dense semantic labels, which limits robustness to emerging land covers and unseen domains~\cite{cao2024open}.

Open-set and open-world perception address unseen categories via unknown rejection, proposal mining, and incremental learning~\cite{Bendale_2016_CVPR,dhamija2020overlooked,joseph2021towards,du2022unknownawareobjectdetectionlearning,Li_2025}. In remote sensing, heterogeneous sensors and evolving taxonomies further complicate these settings. Open-set domain adaptation and large-scale EO Out-of-Distribution (OOD) detection explicitly study such shifts~\cite{zheng2024open,ekim2025distribution}. Open-vocabulary detection and segmentation broadens label spaces through language supervision, but still depends on pre-specified vocabularies and alignment quality~\cite{zhu2024surveyopenvocabularydetectionsegmentation}. In contrast, MVT targets class-agnostic, pixel-accurate region discovery under cross-dataset domain shift. We adopt SAM2 as a class-agnostic front end and apply adaptation to improve prompting robustness on high-resolution remote sensing imagery~\cite{ravi2024sam2segmentimages}.

\subsection{Taxonomy-Grounded Interpretation with MLLMs and Evaluation}
Modern MLLMs~\cite{awadalla2023openflamingo,li2023blip,liu2023visual,bai2025qwen2} enable open-ended naming and rich descriptions, yet can be brittle under open-set conditions: they may over-select from limited candidate sets or hallucinate plausible labels~\cite{kuo2022f,zhou2025led,tu2025ode}. Remote-sensing MLLMs such as GeoChat and RS-LLaVA improve domain awareness~\cite{kuckreja2024geochat,bazi2024rs}, but many studies remain scene-centric and do not explicitly support region-level grounding or standardized, taxonomy-aligned reporting~\cite{ji2021stepwisehierarchicalalignmentnetwork}. Recent open-set RS works begin to leverage MLLMs for unknown discovery and naming by describing or labeling mined proposals~\cite{saini2025advancing,xie2025llama}. MVT advances this direction by using class-agnostic masks as structured grounding evidence. By replacing lexical labels with anonymized identifiers during training, our framework supports a standardized taxonomy interface independent of dataset-specific naming conventions.

For scalable assessment of language outputs, LLM-as-judge protocols provide rubric-based scoring that can correlate with human evaluation~\cite{liu-etal-2023-g}. We adopt GPT-4o for automatic scoring and calibrate it with stratified expert ratings to improve robustness and reproducibility under remote-sensing domain shift~\cite{openai2024gpt4ocard}.

\section{Methodology}\label{sec:method}

\subsection{Data and Lexical-Label-Free Setup}
This study evaluates MVT under cross-dataset domain shift with geometry discovery decoupled from semantic interpretation.
Stage~I adapts a promptable segmenter on OpenEarthMap and tests transfer on LoveDA~\cite{xia2022openearthmapbenchmarkdatasetglobal,wang2022lovedaremotesensinglandcover}.
Stage~II fine-tunes MLLMs in two steps: Step~I uses AID; Step~II adds NWPU-RESISC45, UC Merced, PatternNet, and RSI-CB, and injects SAM2 mask evidence~\cite{xia2017aid,Cheng_2017,yang2010bag,Zhou_2018,li2020rsi,ravi2024sam2segmentimages}. 
All models are evaluated on LoveDA.

To prevent lexical leakage, we remove class-name cues from file paths and metadata by mapping each label to an anonymized ID (e.g., \emph{grassland}$\rightarrow$\emph{category01}) and shuffling all samples into a single directory. 
In total, 70{,}705 samples are used for MLLM tuning (Step~I: 2{,}904; Step~II: 67{,}801).

\subsection{Stage I: Promptable Region Discovery with SAM2}
To enable class-agnostic region discovery, we adopt SAM2~\cite{ravi2024sam2segmentimages} as a class-agnostic, promptable segmentation front end. We compare: (i) off-the-shelf SAM2 pretrained on large-scale generic imagery, and (ii) a domain-adapted SAM2 fine-tuned on OpenEarthMap~\cite{xia2022openearthmapbenchmarkdatasetglobal} using polygon annotations purely as \emph{mask} supervision while discarding category names to avoid semantic leakage. Fine-tuning follows the SAM2 objective of producing high-quality prompt-conditioned masks~\cite{ravi2024sam2segmentimages}.

The research evaluates cross-domain generalization on LoveDA~\cite{wang2022lovedaremotesensinglandcover}, where each tile typically contains multiple co-occurring land-cover types with intricate boundaries. At inference, we apply point prompting to extract pixel-accurate, class-agnostic region masks. These masks serve as structured geometric evidence for Stage~II. For segmentation metrics on LoveDA, we perform IoU matching between predictions and ground truth instances to compute mIoU and mDice.

\subsection{Stage II: Two-Step MLLM Fine-Tuning for Taxonomy Tagging and Mask-Grounded Description} Stage~II performs \emph{tile-level} taxonomy tagging and description generation. Each LoveDA tile is assigned a single dominant land-cover label (the largest-area class in the LoveDA ground truth); SAM2 region masks are used only as grounding cues to support evidence-based reasoning, not as per-region semantic targets. We fine-tune three complementary MLLMs: Qwen2.5-VL-7B~\cite{bai2025qwen2}, Pixtral-12B~\cite{agrawal2024pixtral}, and InternVL3-8B-hf~\cite{zhu2025internvl3}. To keep supervision lexical-label-free, both steps use anonymized numeric pseudo-label IDs as training targets; semantic class names are not used as supervised outputs. We apply LoRA~\cite{hu2022lora,zheng2024llamafactoryunifiedefficientfinetuning} while freezing the vision backbone and multimodal projection layers to preserve pretrained multimodal representations. \paragraph{Step~I (lexical-label-free recognition)} Starting from the pretrained MLLM, Step~I builds basic remote-sensing visual discrimination using the anonymized AID dataset~\cite{xia2017aid}. Training prompts restrict the model to output \emph{only} an anonymized numeric label ID (no free-form explanation) to prevent lexical leakage and force reliance on visual cues (layout, geometry, texture, contrast, and context). We train LoRA with rank $r{=}8$ and $\alpha{=}16$ for 120 steps using a cosine-decay schedule; batch size is 16 with gradient accumulation 8 and context length 8192. \paragraph{Step~II (mask-grounded refinement under a standardized taxonomy interface)} Step~II continues from the Step~I checkpoint by expanding training data with NWPU-RESISC45~\cite{Cheng_2017}, UC Merced~\cite{yang2010bag}, PatternNet~\cite{Zhou_2018}, and RSI-CB~\cite{li2020rsi}, and injecting SAM2-derived region cues as structured prompt inputs. Specifically, for each image/tile we append per-region annotations parsed from SAM2 JSON outputs, including bounding box $\textit{bbox}(x,y,w,h)$, pixel area, and the pixel-level mask encoded as run-length encoding counts~\cite{ravi2024sam2segmentimages}. A standardized land-use taxonomy (Chinese Standard first-level categories)~\cite{chen2007explanation} is included in the prompt as a \emph{reasoning scaffold}, while the supervised target remains strictly the anonymized ID. Step~II uses the same LoRA configuration and frozen vision/projector layers, with optimization adjusted to emphasize subtle, evidence-sensitive distinctions (initial learning rate $1\times10^{-5}$, batch size 4, gradient accumulation 8, 220 steps). At inference and evaluation, we use a fixed prompt that constrains the model to (i) select exactly one Level-1 category from the Chinese-Standard Level-1 list provided in the prompt, (ii) output only the anonymized Level-2 ID, and (iii) generate a mask-grounded description; taxonomy terms are used only as a non-supervised reasoning scaffold.

\subsection{Stage III: Evaluation of Tags and Descriptions}
On LoveDA, we evaluate per-tile Level-1/Level-2 tags and generated descriptions using (i) expert manual scoring and (ii) an LLM-as-judge protocol~\cite{gu2025surveyllmasajudge}. 
Remote-sensing analysts assess tag correctness and description quality, while GPT-4o scores description \emph{naturalness} and \emph{informativeness}, with scores calibrated by stratified expert ratings for reliability~\cite{openai2024gpt4ocard}.

\section{Experiments and Analysis}

\begin{table*}[t]
\centering
\captionsetup{font=scriptsize, skip=2pt}

{\fontsize{7.2}{8.2}\selectfont
\setlength{\tabcolsep}{2.6pt}
\renewcommand{\arraystretch}{0.90}
\begin{tabularx}{\textwidth}{l >{\raggedright\arraybackslash}p{7.2em} >{\raggedright\arraybackslash}p{6.2em} X}
\toprule
\textbf{Model} & \textbf{Level-1} & \textbf{Level-2} & \textbf{Description} \\
\midrule
QW-S & PAPSL & Park and Green Space &
The image shows a park-like area with a road running through it, surrounded by greenery and water bodies. This suggests a recreational or public space designed for the enjoyment and use of the public, which aligns with the characteristics of a park and green space. \\

QW-D & WBHFL & River Surface &
The image shows a water body with a visible river surface, surrounded by land that appears to be developed with roads and some structures, which is consistent with the characteristics of the river surface. \\

IVL-S & WBHFL & Coastal Tidal Flats &
The image shows a coastal area with tidal flats adjacent to a body of water. The presence of a road and some structures near the water indicates a combination of natural and managed land use typical of tidal flat environments. \\

IVL-D & WBHFL & Hydraulic Construction Land &
The image shows a structure adjacent to a water body, likely a \textbf{dam or levee}, indicative of \textbf{hydraulic construction}. The surrounding areas include roads and possibly some vegetation, typical of locations requiring hydraulic management and infrastructure. This setting aligns with features associated with \textbf{water control and management} facilities. \\

PIX-S & Cultivated land & Paddy Field &
The image shows a large expanse of water-filled fields indicative of rice cultivation, which is characteristic of paddy fields. The uniform and structured appearance of the fields, along with the presence of water, aligns with the typical features of paddy fields commonly found in agricultural regions. Additionally, the overall layout and the organized pattern of the fields further support this classification. \\

PIX-D & Cultivated land & Paddy Field &
The image displays a distinct pattern of rectangular plots which are characteristic of organized agricultural land. The presence of water-filled paddies suggests that this is a rice cultivation area, commonly found in regions where paddy fields are prevalent. The uniform coloration and the arrangement further confirm that this is a structured farming system typically used for paddy field. \\
\bottomrule
\end{tabularx}

\caption{Comparison of different MLLMs Outputs in Single-Step (S) and Dual-Step (D) settings based on Figure~\ref{fig:segmentation}. QW: Qwen2.5VL-7B, IVL: InternVL3-8B, PIX: Pixtral 12B. Level-1/2 denote hierarchical land-cover classification. WBHFL: Water Bodies and Hydraulic Facility Land, PAPSL: Public Administration and Public Service Land.}
\label{tab:mlvm_vis}
}

\vspace{0.55em}

\begin{minipage}[t]{\columnwidth}
\vspace{0pt}\centering

{\scriptsize
\setlength{\tabcolsep}{3pt}
\renewcommand{\arraystretch}{0.92}
\begin{tabular}{lcccccc}
\toprule
\textbf{Method} & \textbf{Level-1} & \textbf{Level-1 Acc} & \textbf{Level-2} & \textbf{Level-2 Acc} & \textbf{Descr.} & \textbf{OES} \\
\midrule
QW-S   & 439 & 0.978 & 378 & 0.842 & 384.5 & 8.028 \\
QW-D   & 433 & 0.964 & \textbf{400} & \textbf{0.891} & 406   & \textbf{8.278} \\
IVL-S  & 405 & 0.902 & 331 & 0.737 & 357.5 & 7.306 \\
IVL-D  & \textbf{442} & \textbf{0.984} & 379 & 0.844 & \textbf{408} & 8.212 \\
PIX-S  & 204 & 0.454 & 174 & 0.388 & 185.5 & 3.765 \\
PIX-D  & 247 & 0.550 & 204 & 0.454 & 218   & 4.470 \\
\bottomrule
\end{tabular}
\captionof{table}{Manual scoring results on the LoveDA evaluation subset.}
\label{tab:manual_scores}
}

\vspace{0.55em}

{\scriptsize
\setlength{\tabcolsep}{2pt}
\renewcommand{\arraystretch}{0.92}
\begin{tabular}{lcccccccc}
\toprule
\textbf{Method} & \textbf{Mean} & \textbf{Min} & \textbf{Max} & \textbf{Q1} & \textbf{Q3} & \textbf{Med.} & \textbf{Var.} & \textbf{Std} \\
\midrule
QW-D   & \textbf{92.82} & 69 & 100 & 88.5 & 96.5 & 94.0 & 28.68 & 5.36 \\
IVL-D  & 89.76          & 0  & 100 & 88.0 & 96.5 & 92.5 & 269.97 & 16.43 \\
PIX-D  & 92.47          & 61 & 100 & 88.0 & 96.5 & 92.5 & 30.31 & 5.51 \\
\bottomrule
\end{tabular}
\captionof{table}{GPT-4o naturalness evaluation (scores in [0,100]).}
\label{tab:nat}
}
\end{minipage}
\hfill
\begin{minipage}[t]{\columnwidth}
\vspace{0pt}\centering

{\scriptsize
\setlength{\tabcolsep}{2pt}
\renewcommand{\arraystretch}{0.92}
\begin{tabular}{lcccccccc}
\toprule
\textbf{Method} & \textbf{Mean} & \textbf{Min} & \textbf{Max} & \textbf{Q1} & \textbf{Q3} & \textbf{Med.} & \textbf{Var.} & \textbf{Std} \\
\midrule
QW-D   & \textbf{68.36} & 42.0 & 87.0 & 62.5 & 74.5 & 68.5 & 53.51  & 7.28 \\
IVL-D  & 66.60          & 0.0  & 89.0 & 61.5 & 75.0 & 69.5 & 187.13 & 13.68 \\
PIX-D  & 64.86          & 41.5 & 84.5 & 59.5 & 70.5 & 64.0 & 54.48  & 7.38 \\
\bottomrule
\end{tabular}
\captionof{table}{GPT-4o informativeness evaluation (scores in [0,100]).}
\label{tab:inf}
}

\vspace{0.55em}

\captionsetup{font=scriptsize, skip=2pt}
\noindent
\begin{minipage}[t]{0.73\linewidth}
  \vspace{0pt}\centering
  \begin{subfigure}[t]{0.31\linewidth}
    \centering
    \includegraphics[width=\linewidth]{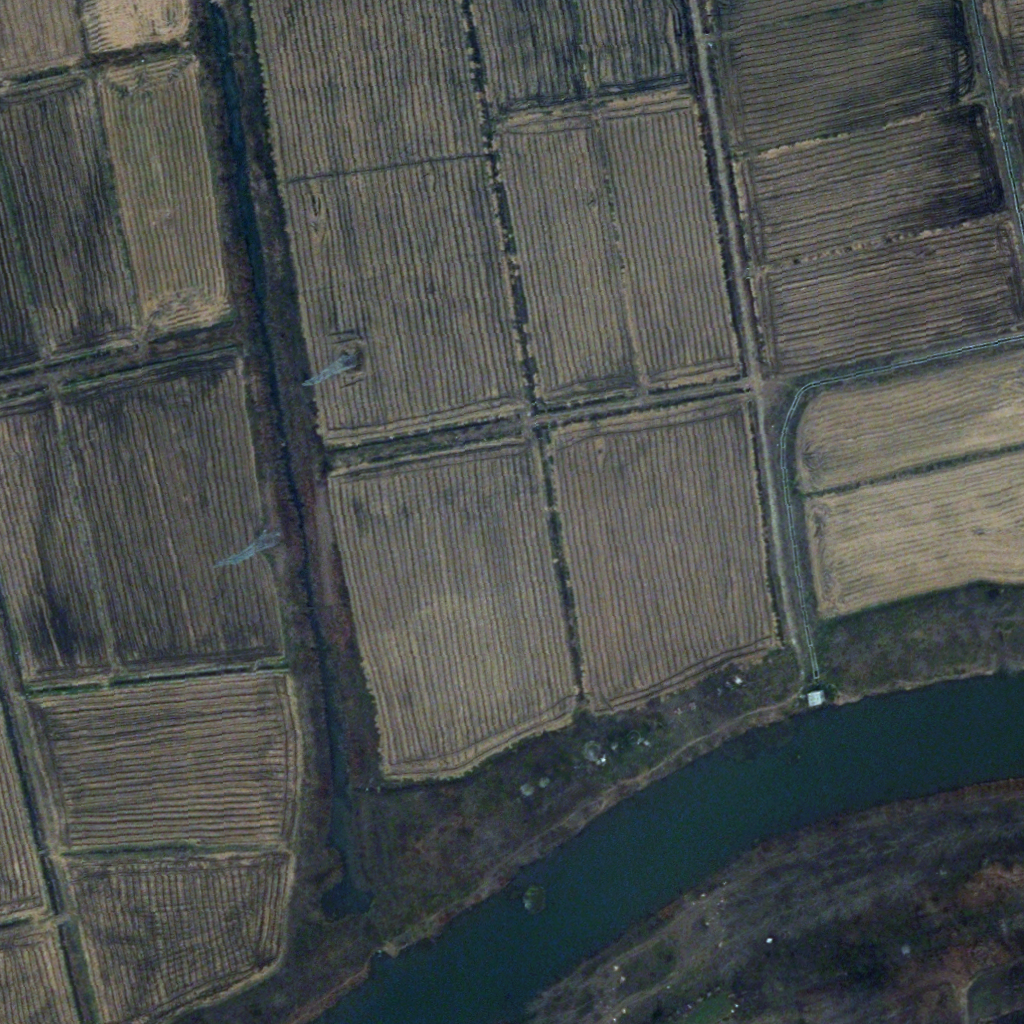}
    \subcaption{}
  \end{subfigure}\hfill
  \begin{subfigure}[t]{0.31\linewidth}
    \centering
    \includegraphics[width=\linewidth]{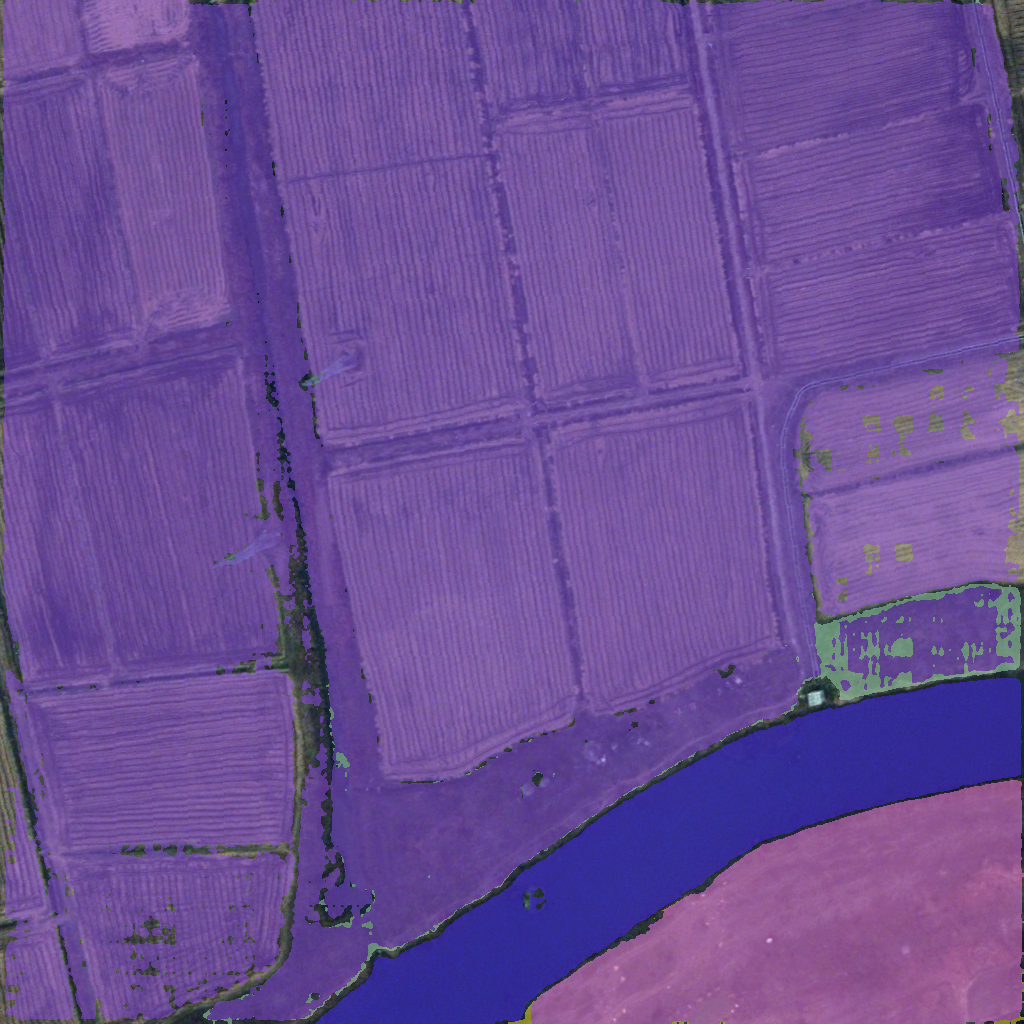}
    \subcaption{}
  \end{subfigure}\hfill
  \begin{subfigure}[t]{0.31\linewidth}
    \centering
    \includegraphics[width=\linewidth]{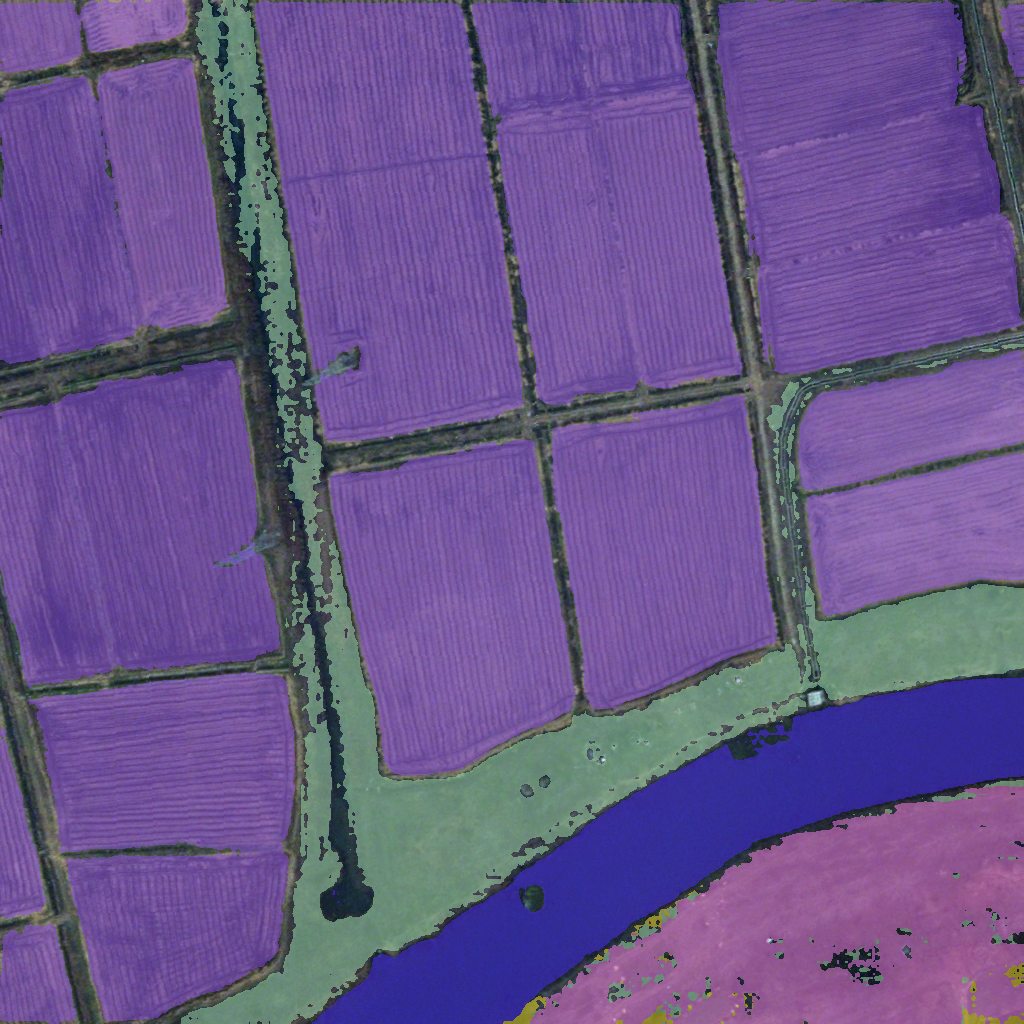}
    \subcaption{}
  \end{subfigure}

  \vspace{-4pt}
  \captionof{figure}{(a) Original Image; (b) Off-the-shelf; (c) Fine-tuned.}
  \label{fig:segmentation}
\end{minipage}
\hfill
\vrule width 0.5pt
\hfill
\begin{minipage}[t]{0.2\linewidth}
  \vspace{0pt}\centering
  \includegraphics[width=\linewidth]{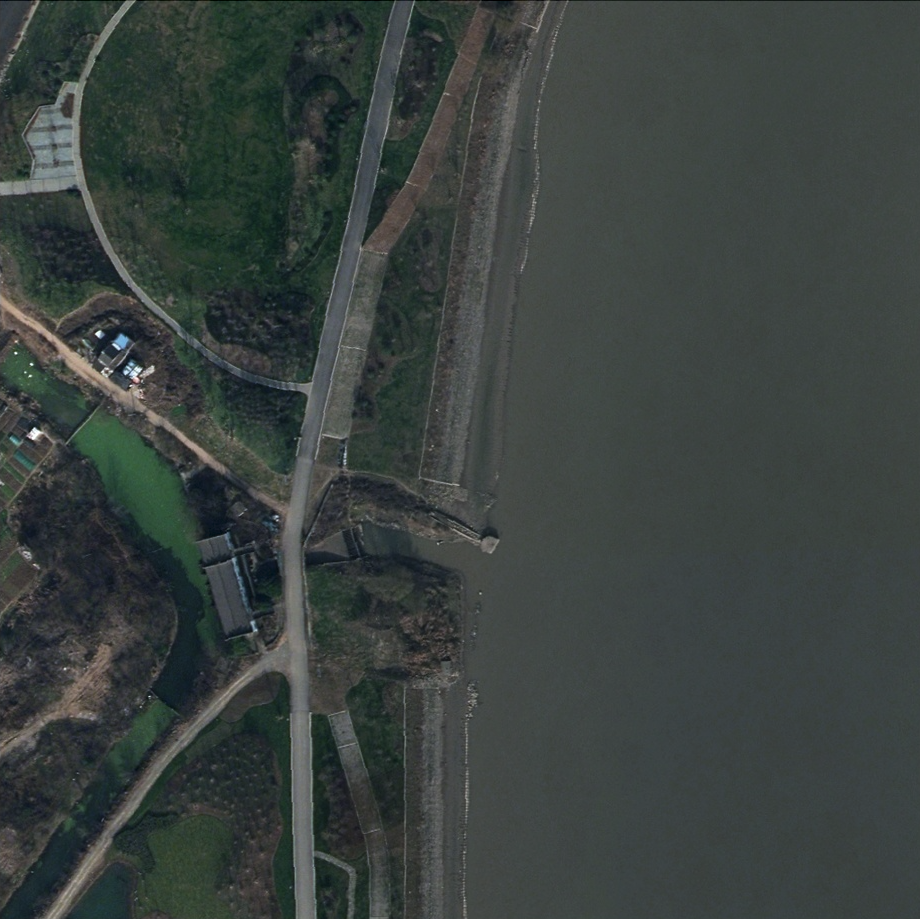}

  \vspace{-2pt}
  \captionof{figure}{Hydraulic Construction Land example.}
  \label{fig:hcl_example}
\end{minipage}

\end{minipage}

\end{table*}

\label{sec:experiments}

\subsection{Environment Setup}

For Stage I, we use the official SAM2 2.1429 Hiera-L checkpoint as the off-the-shelf baseline, finetuning uses AdamW on two 80~GB A100 GPUs (Python~3.10, PyTorch~2.8). For Stage II, we run experiments on two 96~GB NVIDIA H20 GPUs.

\subsection{Segmentation Evaluation}

This study evaluated segmentation quality on the LoveDA dataset, comparing the off-the-shelf SAM2~\cite{ravi2024sam2segmentimages} with a fine-tuned SAM2 variant that is domain-adapted on OpenEarthMap. LoveDA contains multiple co-occurring land-cover types with intricate boundaries~\cite{wang2022lovedaremotesensinglandcover}, posing higher requirements for geometric characterization and generalization ability. We report mean Intersection over Union (mIoU) and mean Dice score (mDice)~\cite{qiu2025noiseconsistentsiamesediffusionmedicalimage}.

For each annotated instance $i\in\mathcal{I}$, with predicted mask $P_i$ and ground-truth mask $G_i$,
{\scriptsize
\begin{align}
\mathrm{IoU}_i &= \frac{|P_i\cap G_i|}{|P_i\cup G_i|},\quad
\mathrm{Dice}_i = \frac{2|P_i\cap G_i|}{|P_i|+|G_i|}.
\end{align}
We report dataset-level instance means:
\begin{align}
\mathrm{mIoU} &= \frac{1}{|\mathcal{I}|}\sum_{i\in\mathcal{I}}\mathrm{IoU}_i,\quad
\mathrm{mDice} = \frac{1}{|\mathcal{I}|}\sum_{i\in\mathcal{I}}\mathrm{Dice}_i.
\end{align}
}

On LoveDA, SAM2 fine-tuned on OpenEarthMap improves instance-level mIoU/mDice to 0.4444/0.5546 vs. 0.3745/0.4854 off-the-shelf, and yields finer boundaries in complex urban–rural scenes (Fig.~\ref{fig:segmentation}).

\subsection{MLLM Evaluation}
We evaluate MLLMs on LoveDA~\cite{wang2022lovedaremotesensinglandcover} to assess (i) tile-level Level-1/Level-2 tagging accuracy and (ii) description quality in our class-agnostic discovery-and-interpretation setting. We conduct a qualitative ablation by comparing Single-Step (S) baseline—finetuned exclusively on Step~II—with our proposed Dual-Step (D) pipeline, which follows the complete two-step schedule.

\subsubsection{Visualization of Semantic Tagging}
Table~\ref{tab:mlvm_vis} compares Single-Step vs.\ Dual-Step outputs. InternVL3 benefits most from Dual-Step finetuning: IVL-D correctly identifies \emph{Hydraulic Construction Land} by grounding it in man-made infrastructures as dams and levees, whereas IVL-S identifies it as natural coastal patterns in tidal flats. Pixtral (PIX-S/D) is consistently reliable on \emph{Paddy Field}. Qwen2.5VL shows clear refinement with Dual-Step tuning: QW-S misclassifies the scene as a public park, while QW-D shifts to \emph{Water Bodies} (River Surface). Overall, Dual-Step tuning improves discrimination between subtle man-made hydraulic infrastructures and visually similar natural water or coastal scenes.

These qualitative gains match the intended step-wise roles: Step~I builds coarse RS discrimination under lexical-label-free, ID-only supervision; Step~II injects region-level geometric evidence (SAM2 masks, bounding boxes, and areas) to encourage explicit grounding and fine-grained subtype reasoning. Consistently, improvements are more pronounced for Level-2 tagging and description quality than for Level-1 categorization (Table~\ref{tab:manual_scores}).

\subsubsection{Manual Scoring}

Manual scoring is designed to rigorously assess semantic tagging accuracy at the scene (per-tile) level. We randomly select 25\% (449 images) of the LoveDA Dataset for evaluation. The evaluation metric incorporates three dimensions: first-level tagging, second-level tagging, and description quality. For each sample, Level-1 and Level-2 categories are scored binarily (0/1), while the description is graded on a three-level scale (0 for incorrect, 0.5 for partially correct, and 1 for fully correct), with a per-sample maximum of 3 points. The overall evaluation score (OES; full score is 9) is

{\scriptsize
\begin{align}
\text{OES} = \frac{\text{Level-1} + \text{Level-2} + \text{Description}}{\text{Number of Samples}} \times 3.
\end{align}
}

Results in Table~\ref{tab:manual_scores} reveal a clear distinction between single and dual-step finetuned models. Dual-Step finetuned Qwen2.5VL and InternVL3 achieve the highest overall scores (8.278 and 8.212), significantly outperforming their single-finetuned counterparts. Pixtral shows the weakest tagging accuracy even after dual-step finetuning.


\subsubsection{LLM-as-Judge with GPT-4o}

To systematically assess the linguistic quality of generated descriptions, we adopt an LLM judge protocol~\cite{gu2025surveyllmasajudge} using GPT-4o~\cite{openai2024gpt4ocard}. Descriptions are evaluated along two axes: \emph{naturalness} and \emph{informativeness}. Naturalness is scored via five weighted sub-modules: \textbf{grammar \& syntax} (0.25), \textbf{discourse coherence \& flow} (0.25), \textbf{lexical naturalness \& idiomaticity} (0.20), \textbf{style \& register appropriateness} (0.15), and \textbf{human-likeness vs.\ Machine ``tells''} (0.15). Informativeness is scored via \textbf{coverage of key facets} (0.25), \textbf{specificity \& quantification} (0.25), \textbf{concreteness \& observability} (0.20), \textbf{context, constraints \& relations} (0.20), and \textbf{relevance \& non-redundancy} (0.10). To ensure deterministic and objective scoring, we set the sampling temperature to $0$ and utilize structured prompting to enforce single-line fixed numerical output format. The weighting follows established linguistic quality frameworks and manual-evaluation practice~\cite{lommel2013multidimensional}.

The per-sample score, scaled to $[0,100]$, is computed as
{\scriptsize
\begin{align}
\text{Score}_{\text{total}} = 100 \times \sum_{i=1}^{5} w_i \left( \frac{s_i}{5} \right),
\end{align}
}

Tables~\ref{tab:nat} and~\ref{tab:inf} report descriptive statistics for naturalness and informativeness. For naturalness, Qwen obtains the highest mean (92.82), closely followed by Pixtral and InternVL3. For informativeness, Qwen again leads (68.36), with InternVL3 and Pixtral trailing. Combining manual accuracy and GPT-4o-based naturalness and informativeness, Qwen and InternVL3 emerge as the most effective models for taxonomy-grounded RS tagging and description in our discovery setting, while Pixtral shows limited semantic tagging capability despite reasonable linguistic fluency.



\section{Conclusion}
\label{sec:conclusion}

In this work, we presented MVT, a geometry-first framework designed to address the challenges of land-cover understanding under cross-dataset domain shift. By decoupling geometric discovery from semantic interpretation, MVT effectively leverages domain-adapted SAM2 for boundary-faithful mask extraction and employs a dual-step MLLM fine-tuning strategy to generate taxonomy-aligned tags and grounded descriptions. Crucially, our ID-based supervision strategy bypasses reliance on dataset-specific lexical labels, enabling a flexible, class-agnostic interface. Experimental results on the LoveDA dataset demonstrate that MVT significantly outperforms single-step baselines in both tagging accuracy and descriptive richness. Future work will extend this framework to open-vocabulary scenarios and evaluate its robustness against broader withheld-class protocols.

\small
\bibliographystyle{IEEEtranN}
\bibliography{references}

\end{document}